\documentclass[sigconf]{acmart}

\usepackage{makecell}
\newcommand{\pluseq}{\mathrel{+}=}
\def\ie{{\em i.e. }}
\def\eg{{\em e.g., }}

\usepackage{balance}

\def\BibTeX{{\rm B\kern-.05em{\sc i\kern-.025em b}\kern-.08emT\kern-.1667em\lower.7ex\hbox{E}\kern-.125emX}}

\copyrightyear{2019}
\acmYear{2019}
\acmConference[MM '19]{Proceedings of the 27th ACM International Conference on Multimedia}{October 21--25, 2019}{Nice, France}
\acmBooktitle{Proceedings of the 27th ACM International Conference on Multimedia (MM '19), October 21--25, 2019, Nice, France}
\acmPrice{15.00}
\acmDOI{10.1145/3343031.3350534}
\acmISBN{978-1-4503-6889-6/19/10}

\begin{document}

\fancyhead{}

\title{daBNN: A Super Fast Inference Framework for Binary Neural Networks on ARM devices}


\author{Jianhao Zhang, Yingwei Pan, Ting Yao, He Zhao and Tao Mei}
\affiliation{%
  \institution{JD AI Research, Beijing, China}}
\email{{daquexian566,panyw.ustc,tingyao.ustc}@gmail.com;zhaohe5@jd.com;tmei@jd.com}

%
\renewcommand{\shortauthors}{Zhang and Pan, et al.}

%
\begin{abstract}
      It is always well believed that Binary Neural Networks (BNNs) could drastically accelerate the inference efficiency by replacing the arithmetic operations in float-valued Deep Neural Networks (DNNs) with bit-wise operations. Nevertheless, there has not been open-source implementation in support of this idea on low-end ARM devices (e.g., mobile phones and embedded devices). In this work, we propose daBNN --- a super fast inference framework that implements BNNs on ARM devices. Several speed-up and memory refinement strategies for bit-packing, binarized convolution, and memory layout are uniquely devised to enhance inference efficiency. Compared to the recent open-source BNN inference framework, BMXNet, our daBNN is $7\times$$\sim$$23\times$ faster on a single binary convolution, and about $6\times$ faster on Bi-Real Net 18 (a BNN variant of ResNet-18). The daBNN is a BSD-licensed inference framework, and its source code, sample projects and pre-trained models are available on-line: https://github.com/JDAI-CV/dabnn.
\end{abstract}

%
%

\ccsdesc[500]{Software and its engineering~Software libraries and repositories}
\ccsdesc[300]{Computer systems organization~Neural networks}

\keywords{Open Source; Binary Neural Networks; Machine Learning}

%

%
\maketitle

\section{Introduction}

The advances in Deep Neural Networks (DNNs) have substantially pushed the limits and reached new state-of-the-arts of technologies in multimedia and computer vision areas. These advancements rely heavily on the requirements to have high-performance computational accelerator---GPUs. Nevertheless, there has been exponential growth in DNN-based apps available on-line for low-end ARM devices (e.g., mobile phones) recently. For instance, in the month of Sep. 2018, Android users downloaded 221 DNN-based apps for around 13 million times from the official Google Play market and the number of DNN-based apps increased by 27\% over 3 months \cite{xu2019first}. Meanwhile, most major vendors developed DNN inference frameworks tailored to ARM devices, e.g., TensorFlow Lite \cite{tensorflowlite} from Google and Caffe2 \cite{caffe2} from Facebook, which quickly gains popularity for their stronger privacy protection and lower cost for data transmission. However, the inference efficiency of DNNs on ARM devices is often limited with relatively small memory storage and inferior computing power of mobile phones or embedded devices. This adversely hinders the deployment of heavy DNNs on ARM devices. One feasible way to alleviate this problem is to utilize Binary Neural Networks (BNNs), which quantize the activations and weights of DNNs to 1-bit, and lead a significant speed-up over the full precision\footnote{Full precision indicates 32-bit float (the normal precision in DNNs).} counterpart with efficient bit-wise operations.

In the literature, there are several inference frameworks for BNNs: BitStream \cite{Zhao:2018:BEC:3240508.3240673}, BitFlow \cite{hu2018bitflow}, and BMXNet \cite{bmxnet}. Among them, BitStream and BitFlow are not open-source and not available to public. To our best knowledge, the only one open-source BNN inference framework is BMXNet \cite{bmxnet}, which is found to be even slower than full-precision TensorFlow Lite in our experiments. Such facts motivate and highlight the explorations of an open-source BNN inference framework highly optimized for ARM devices.

To address such problems, we present daBNN, a super fast inference framework that implements BNNs on ARM devices with several uniquely devised technologies for speeding up the inference. In particular, an upgraded bit-packing scheme is adopted to pack multiple elements simultaneously, which improves the speed of naive sequential method by about 4$\times$. Moreover, our daBNN capitalizes on ``binary direct convolution'' to squeeze the cost of additional instructions in binary convolution, and meanwhile, a novel memory layout is leveraged to reduce memory access. daBNN is written in C++ and ARM assembly, and provides Java binding and Android package. Experiments demonstrate that our daBNN is characterized with extremely fast speed. Specifically, compared to BMXNet, our daBNN is 7$\times$-23$\times$ faster on a single binary convolution, and about 6$\times$ faster on Bi-Real Net 18 \cite{liu2018bi} (a BNN variant of ResNet-18 \cite{He_2016}). When comparing to full-precision TensorFlow Lite, our daBNN is 8$\times$-10$\times$ faster on a single binary convolution, and about 3$\times$ faster on Bi-Real Net 18. We believe our daBNN will offer a fertile ground for deploying BNNs in industry and designing novel BNN structures in academia.

\section{daBNN}

In this section, we present the detailed implementations in our daBNN, the comparison to existing software, and the potentiality to help the architecture design of BNNs.

\subsection{Implementation Details}

\subsubsection{Bit-packing}

\begin{figure}
    \centering
    \includegraphics[scale=0.43]{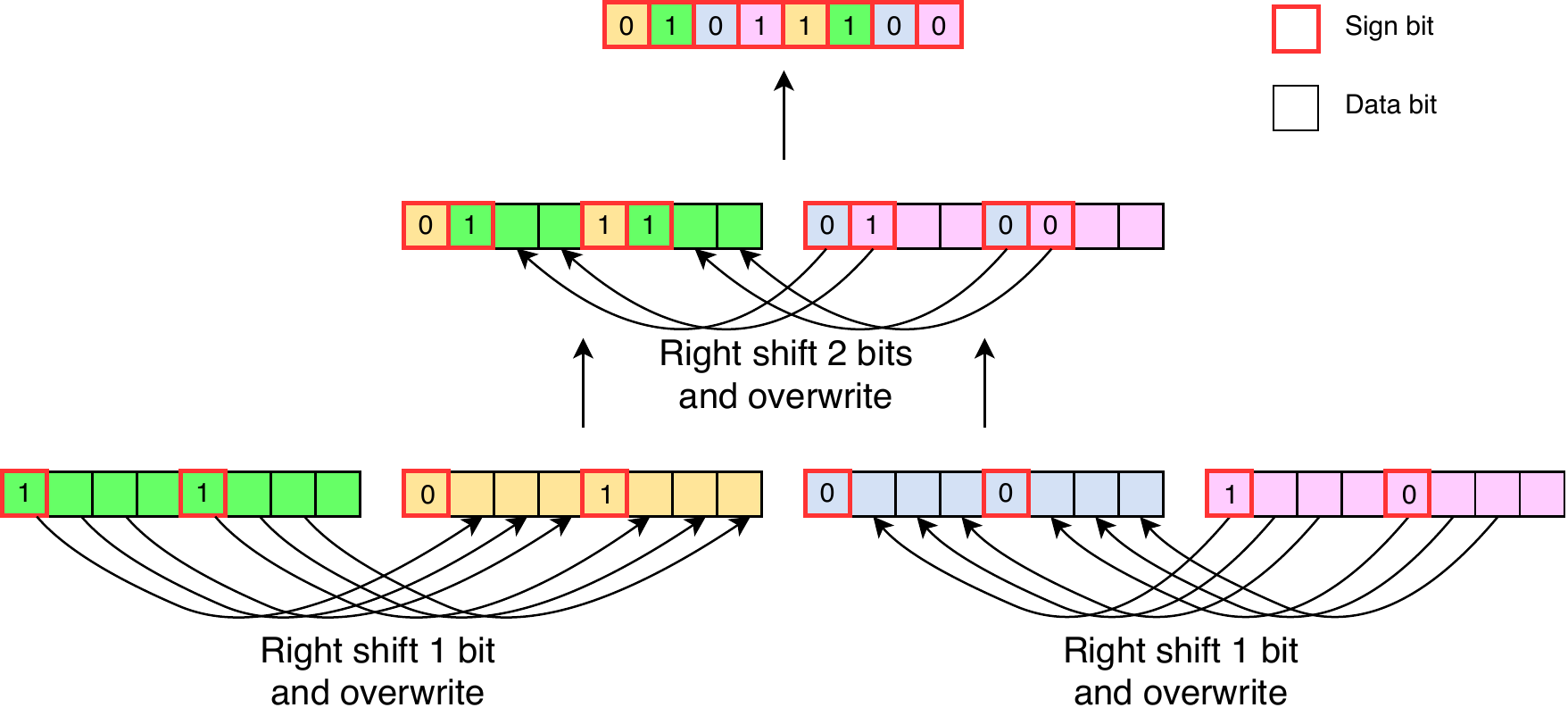}
    \vspace{-0.2in}
    \caption{Our upgraded bit-packing scheme by packing eight 4-bit elements (four 8-bit vectors) into an 8-bit vector.}
    \label{fig:bitpack}
\end{figure}

\begin{figure}
    \centering
    \includegraphics[scale=0.55]{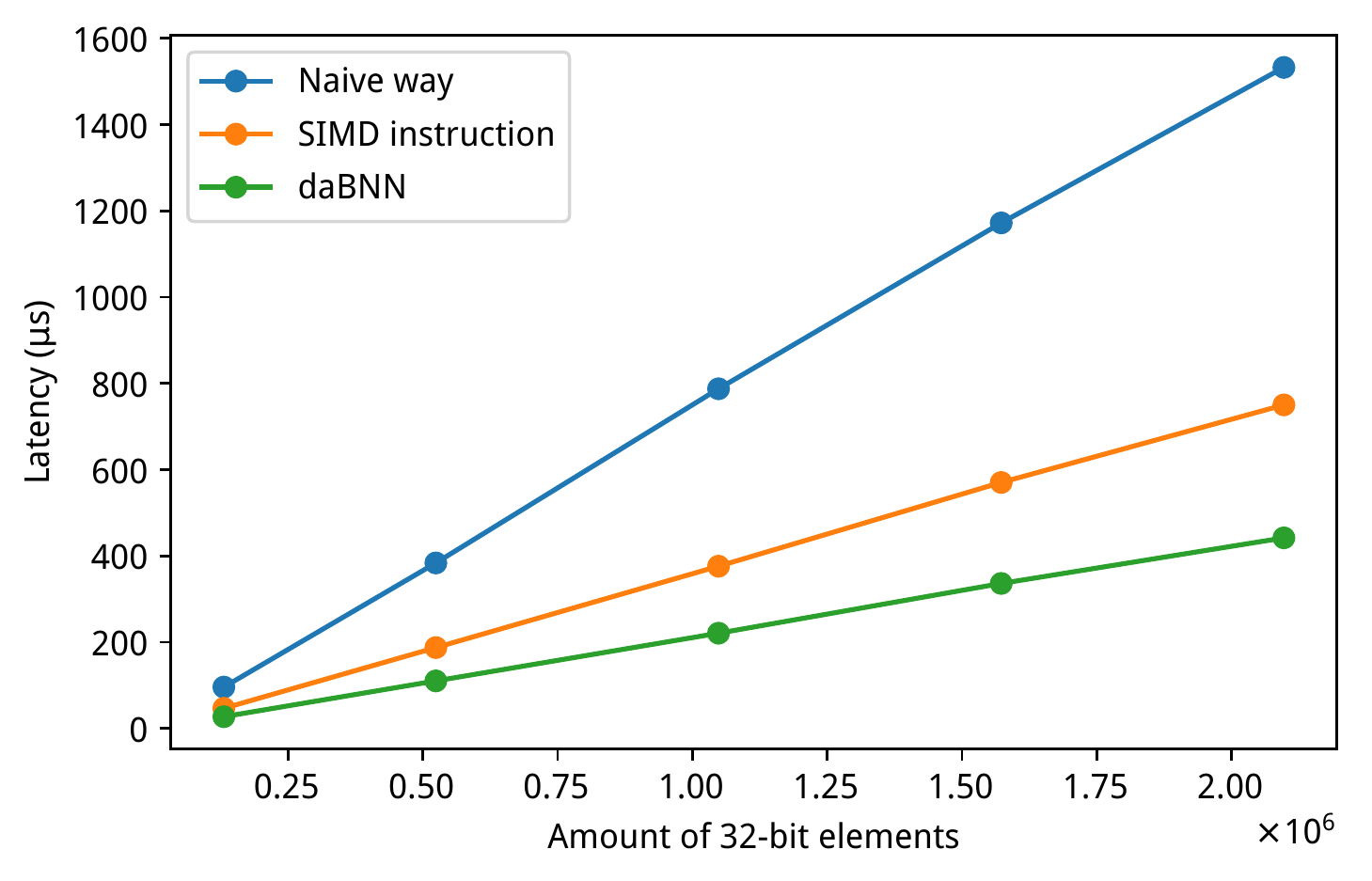}
    \vspace{-0.2in}
    \caption{Latency comparison between our upgrade bit-packing scheme and other methods.}
    \label{fig:diff_bitpack}
    \vspace{-0.2in}
\end{figure}

Bit-packing is a common scheme for BNNs by binarizing $N$ (\eg 128) elements into 1-bit (i.e., 1 or 0), and then packing them into an $N$-bit vector. As such, xnor can be directly performed between these binarized vectors. Previous works \cite{hu2018bitflow}\cite{bmxnet}\cite{Zhao:2018:BEC:3240508.3240673} often perform bit-packing in a naive way. For example, \cite{hu2018bitflow} compares every 32-bit element with zero and sets the corresponding bit sequentially. Unlike them, we directly utilize the existing sign bit in int32 and IEEE 754 float numbers without the need for additional comparison with zero. Moreover, the ``right-shift-and-overwrite'' SIMD (single instruction, multiple data) instruction is adopted here, which enables the simultaneous gathering of multiple sign bits. Such instruction takes three operands: two vectors $\alpha$ and $\beta$ (both of which contain several $M$-bit elements) and a scalar $k$. It performs the right shift operation over every $M$-bit element in vector $\alpha$ by $k$ bits, and overwrites the rightmost $M-k$ bits of each $M$-bit element in vector $\beta$. In addition, we further upgrade bit-packing by scattering these instructions on different registers to avoid write-after-write data hazard, as shown in Figure \ref{fig:bitpack}. The experimental results in Figure \ref{fig:diff_bitpack} illustrate that our upgraded bit-packing scheme is $\sim$4$\times$ as fast as the naive way. Note that our method is also compatible with fused-BN-binarization layer in \cite{Zhao:2018:BEC:3240508.3240673}.

\subsubsection{Binary Direct Convolution}
SGEMM (Single float GEneral Matrix Multiplication) is a widely adopted approach to implement float convolutions in various high-performance scientific programs. In the context of BNNs, an alternative operation to SGEMM is BGEMM, which performs binary matrix multiplication for binary convolution. In addition to the common multiplication and add operations, BGEMM includes extra operations that count how many 1s are in a vector. Specifically, we denote $U^{M \times N}$ as the space of matrices with dimension $M \times N$ and each element of it is a bit-packed vector. Given two matrices (i.e., $ \boldsymbol{A} \in U^{M \times K}$ and $\boldsymbol{B} \in U^{K \times N}$), $\boldsymbol{C} \in \mathbb{N}^{M \times N}$ ($\mathbb{N}$ represents the set of non-negative integers), $\boldsymbol{C} = BGEMM(\boldsymbol{A}, \boldsymbol{B})$ is measured as:
\begin{equation} \label{binary_multiply_add}
    C_{i,j} = \sum\nolimits_{k} bitcount(xnor(\Vec{A_{i,k}}, \Vec{B_{k,j}})),
\end{equation}
where $\Vec{A_{i,k}}$ and $\Vec{B_{k,j}}$ denotes each element in $ \boldsymbol{A}$ and $\boldsymbol{B}$. In SGEMM, to amortize the cost of loading memory, $\boldsymbol{C}$ is often calculated as
\begin{equation} \label{mul_col_row_0}
    \boldsymbol{C^{k}} = \boldsymbol{m^{k}}\boldsymbol{n^{k}},
\end{equation}
\begin{equation} \label{mul_col_row}
    \boldsymbol{C} \pluseq \boldsymbol{C^{k}},
\end{equation}
where $\boldsymbol{m^{k}}$ is the $k_{th}$ column of $\boldsymbol{A}$ and $\boldsymbol{n^{k}}$ is the $k_{th}$ row of $\boldsymbol{B}$.

We argue that this way is sub-optimal for BGEMM especially on ARM devices. In particular, on ARMv8 (the 64-bit ARM architecture) devices, the operation of bitcount contains two instructions: ``cnt'' and ``addv''. ``cnt'' takes an $N$-byte vector $\alpha$ as input and outputs an $N$-byte vector $\beta$, which $\beta_{i} = the\_number\_of\_1s(\alpha_{i})$ where $\alpha_{i}$ and $\beta_{i}$ are the $i_{th}$ byte of $\alpha$ and $\beta$ respectively. ``addv'' sums up all bytes in a vector and outputs the aggregated scalar. Eq. \ref{mul_col_row} is then expanded as:
\begin{equation} \label{mul_col_row_expanded}
    C_{i,j} \pluseq addv(cnt(xnor(\Vec{m^{k}_{i}}, \Vec{n^{k}_{j}}))).
\end{equation}

Thus, Eq. \ref{mul_col_row_expanded} shows that the operation of binary multiply-addition on ARMv8 devices consists of four instructions: xnor, cnt, addv, and addition. Moreover, on ARMv7 (the 32-bit ARM architecture) devices, there is even no ``addv'' instruction and $\lceil \log_{2}N \rceil$ instructions are needed to sum up all bytes in an $N$-byte vector, so the operation of binary multiply-addition consists of $\lceil \log_{2}N \rceil+3$ instructions on these devices. To improve the efficiency of this operation, we re-arrange the calculation order and calculate $\boldsymbol{C}=BGEMM(\boldsymbol{A},\boldsymbol{B})$ as the multiplication of a row vector $\boldsymbol{p} \in U^{1 \times N}$ and $\boldsymbol{q} \in U^{M \times 1}$:
\begin{equation}
    C_{i,j} = \boldsymbol{p^{i}}\boldsymbol{q^{j}},
\end{equation}
where $\boldsymbol{p^{i}}$ is the $i_{th}$ row of $\boldsymbol{A}$ and $\boldsymbol{q^{j}}$ is the $j_{th}$ column of $\boldsymbol{B}$.

In this way, the cost of ``addv'' instructions can be mostly squeezed by summing up the results of ``cnt'' in advance:
\begin{equation} \label{binary_conv_armv8}
    \Vec{C_{i,j}} = \sum\nolimits_{k} cnt(xnor(\Vec{A_{i,k}}, \Vec{B_{k,j}})),
\end{equation}
\begin{equation} \label{binary_conv_armv8_2}
    C_{i,j} = addv(\Vec{C_{i,j}}).
\end{equation}

Please note that the same transformation can not be employed in Eq. \ref{mul_col_row_expanded} because $\boldsymbol{C}$ is stored as 32-bit integers to save the valuable registers. Therefore in Eq. \ref{mul_col_row_expanded}, we have to utilize ``addv'' to reduce the vector into an integer before every instruction of ``addition''. Taking a close look on Eq. \ref{binary_conv_armv8} and \ref{binary_conv_armv8_2}, we can observe some interesting connections between them and the operation of convolution. Specifically, if we treat $\boldsymbol{A} \in U^{M \times K}$ and $\boldsymbol{B} \in U^{K \times N}$ as the weight and the im2col-ed input ($M$: the number of output channels, $N$: output height $\times$ output width, and $K$: the number of bit-packed vectors in a weight filter), Eq. \ref{binary_conv_armv8} and \ref{binary_conv_armv8_2} can be directly interpreted as the definition of convolution. As such, the refined operation of binary convolution is dubbed as ``binary direct convolution''.

\begin{figure}
    \centering
    \includegraphics[scale=0.4]{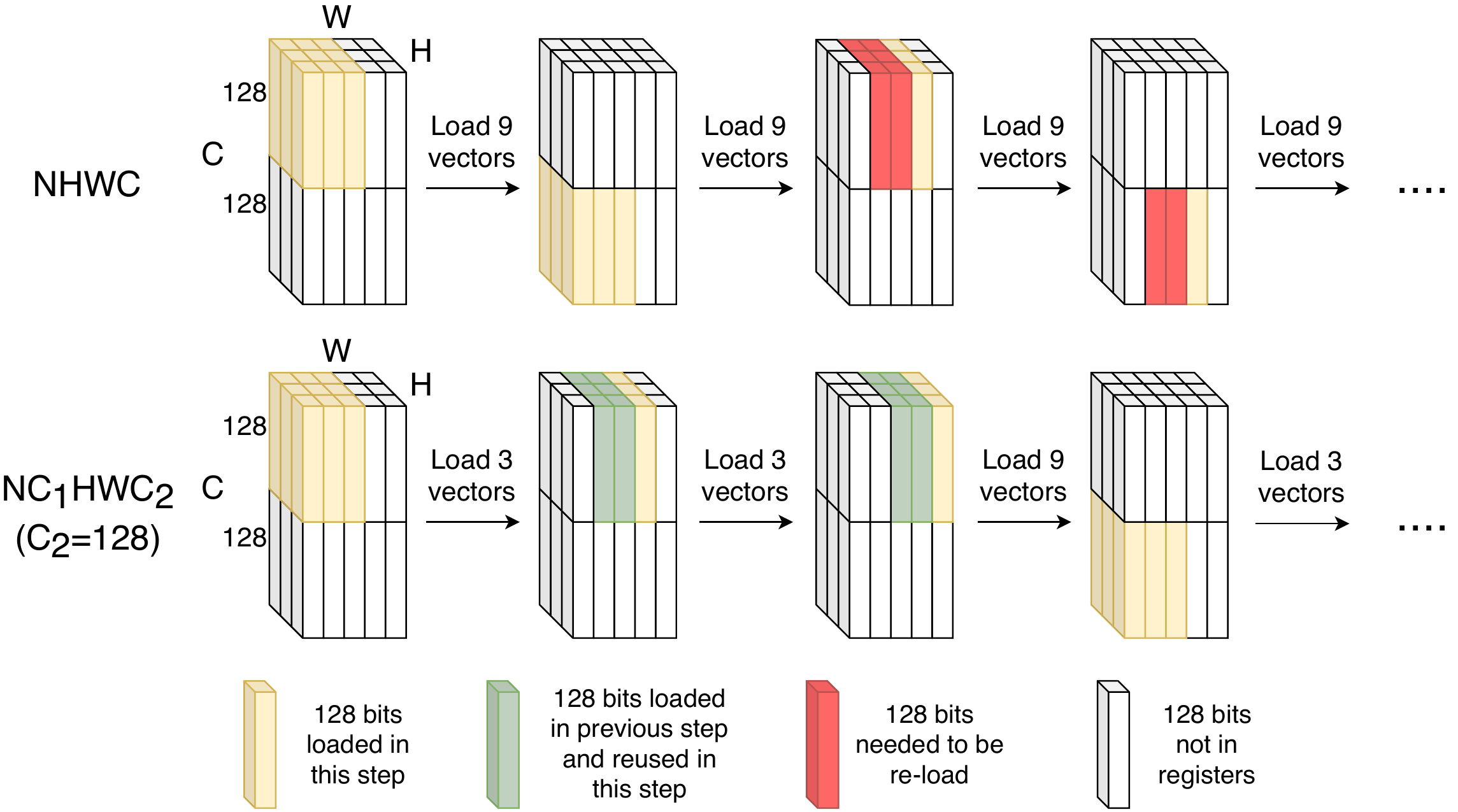}
    \caption{The register reusing of a 3x3 convolution in NC\textsubscript{1}HWC\textsubscript{2} memory layout. It shows that this new layout can save about $2/3$ memory loading.}
    \label{fig:nc1hwc2}
    \vspace{-0.2in}
\end{figure}

\begin{table*}[]
    \centering
    \begin{tabular}{c|c|c|c|c|c|c}
        \hline
                    & Open Source   & Language      & Method & Memory Layout      & Speed on Google Pixel 1  & Model Format  \\
        \hline
        BMXNet      & \textbf{Yes}           & Intrinsics    & BGEMM  & NHWC          & \makecell{3$\times$ faster than Caffe \\ Slower than TF Lite} & \makecell{MXNet with \\ custom operators}         \\
        \hline
        BitStream   & Unavailable & -             & BGEMM  & NHWC          & -          & -       \\
        \hline
        BitFlow     & Unavailable & -             & \multicolumn{3}{c|}{No ARM implementation}          & -       \\
        \hline
        daBNN       & \textbf{Yes}           & \textbf{Assembly}      & \makecell{\textbf{Binary Direct Conv} \\ \textbf{and BGEMM}} & \textbf{NC\textsubscript{1}HWC\textsubscript{2}} & \makecell{\textbf{20$\times$ faster than Caffe} \\ \textbf{6.5$\times$ faster than TF Lite}}          & \textbf{ONNX}          \\
        \hline
    \end{tabular}
    \caption{Comparison to existing software. Only BMXNet and daBNN is open-source, while other software (i.e., BitStream and BitFlow) are unavailable to public. BitFlow does not report its implementation and speed on ARM devices. BMXNet depends on custom MXNet operators like ``QConvolution'', so models trained in other frameworks cannot be deployed on it easily. By contrast, our daBNN is constructed based on standard ONNX operators (Sign and Convolution) to ensure the interoperability.}
    \label{tab:comparison}
    \vspace{-0.2in}
\end{table*}

\begin{figure}
    \centering
    \vspace{-0.1in}
    \includegraphics[scale=0.5]{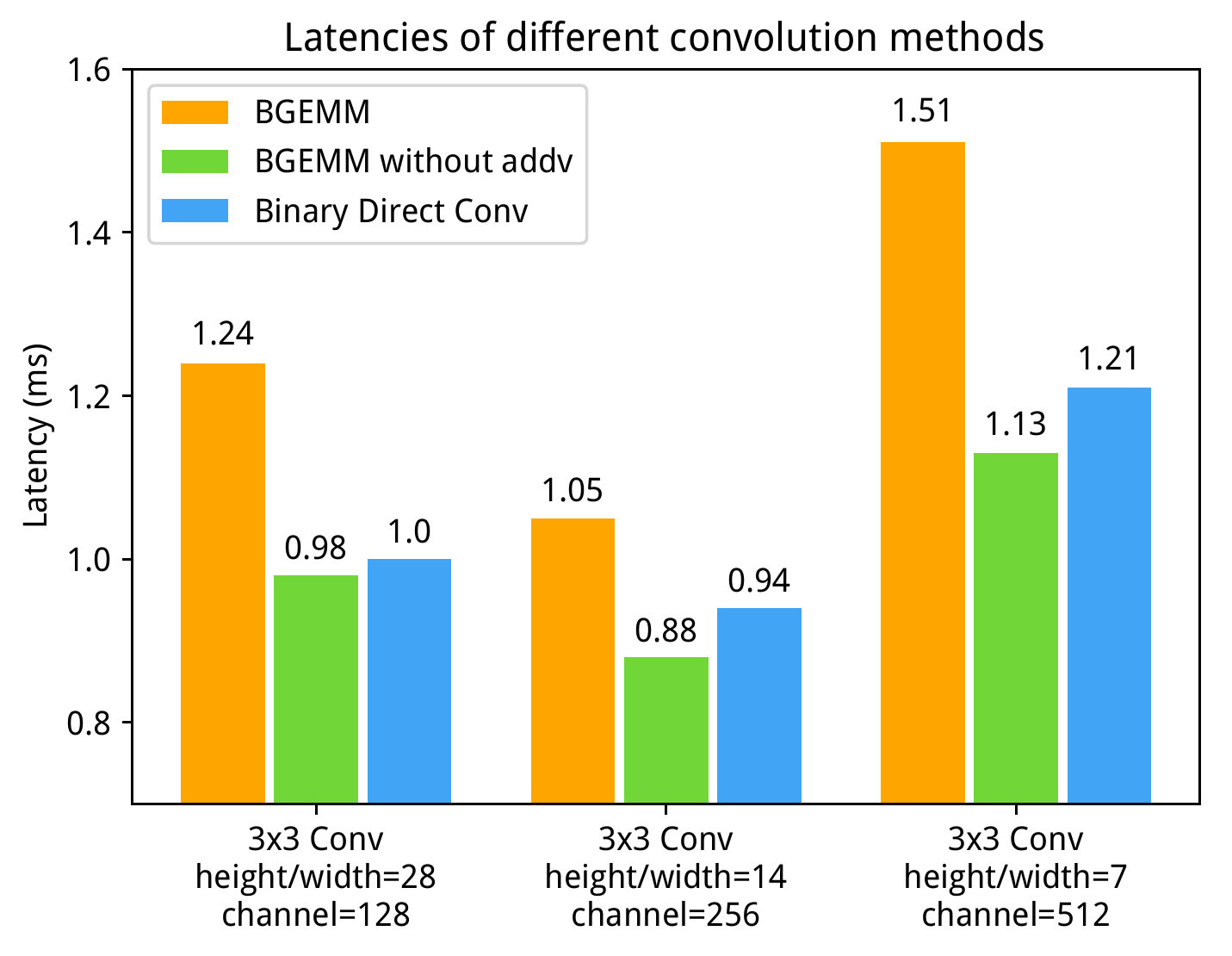}
    \vspace{-0.1in}
    \caption{Latency comparison between different convolution methods. ``BGEMM without addv'' denotes an abnormal implementation of BGEMM by removing the ``addv'' instructions. It clearly shows that the ``addv'' instructions make BGEMM slower than our binary direct convolution.}
    \label{fig:diff_conv_method}
    \vspace{-0.1in}
\end{figure}

Though our binary direct convolution enhances the efficiency by excluding most addv instructions, it results in more cost on memory access. To compensate the increase on memory access, a novel memory layout, NC\textsubscript{1}HWC\textsubscript{2} is devised to leverage the spatial redundancy of convolutions, where C\textsubscript{1} = C/C\textsubscript{2} and C\textsubscript{2} is the length of a register (\ie 128 on ARM devices). NC\textsubscript{1}HWC\textsubscript{2} can be regarded as a refinement of NHWC by splitting NHWC layout into several groups, where each group has C\textsubscript{2} bits. As such, each register holds all channels in a group. A diagram of NC\textsubscript{1}HWC\textsubscript{2} is illustrated in Figure \ref{fig:nc1hwc2}, which indicates that $2/3$ registers of previous location are reused in this refined memory layout (i.e., $2/3$ memory access is saved). The experimental results in Figure \ref{fig:diff_conv_method} clearly show that the extra ``addv'' instructions make BGEMM slow, and our binary direct convolution with NC\textsubscript{1}HWC\textsubscript{2} is faster than BGEMM. Note that while we only present the implementation details on ARMv8 here, the proposed binary direct convolution and NC\textsubscript{1}HWC\textsubscript{2} memory layout are also compatible and effective on ARMv7.

\subsection{Comparison to Existing Software}

We compare our daBNN and other BNN inference frameworks in Table \ref{tab:comparison}, and summarize the following two key differences:

(1) daBNN is open-source. Although BitStream and BitFlow are claimed to be fast, neither of them is available on-line.

\begin{figure}
    \centering
    \includegraphics[scale=0.5]{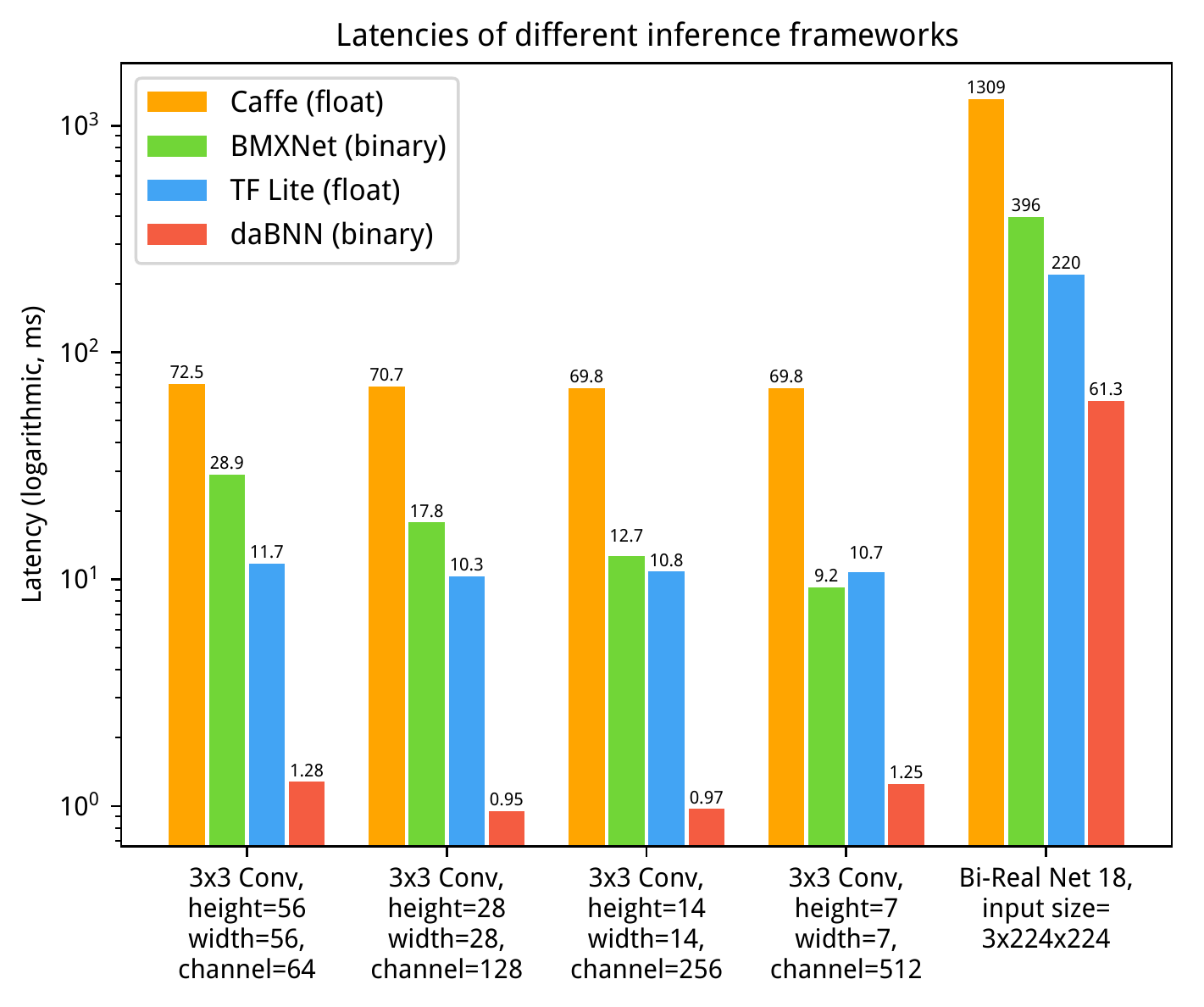}
    \vspace{-0.15in}
    \caption{Latency comparison between different inference frameworks on various convolutions and Bi-Real Net 18. The latency is measured on Google Pixel 1 (single thread). daBNN achieves the fastest inference speed, while BMXNet is even slower than the full precision TensorFlow Lite.}
    \label{fig:comparison}
    \vspace{-0.05in}
\end{figure}

(2) The speed of daBNN is super fast. To our best knowledge, daBNN is the first BNN implementation that is compared with modern mobile inference framework (e.g., TensorFlow Lite) rather than Caffe \cite{Jia:2014:CCA:2647868.2654889} or OpenBLAS. Figure \ref{fig:comparison} compares the latency between our daBNN and existing software (i.e., TensorFlow Lite, BMXNet, and Caffe). Compared to TensorFlow Lite, our daBNN is 8$\times$-10$\times$ faster on a single binary convolution, and about 3$\times$ faster on Bi-Real Net 18. Note that the only one existing open-source BNN implementation, BMXNet, is even slower than TensorFlow Lite on Bi-Real Net 18 and several convolutions.

\begin{figure}
    \centering
    \includegraphics[scale=0.8]{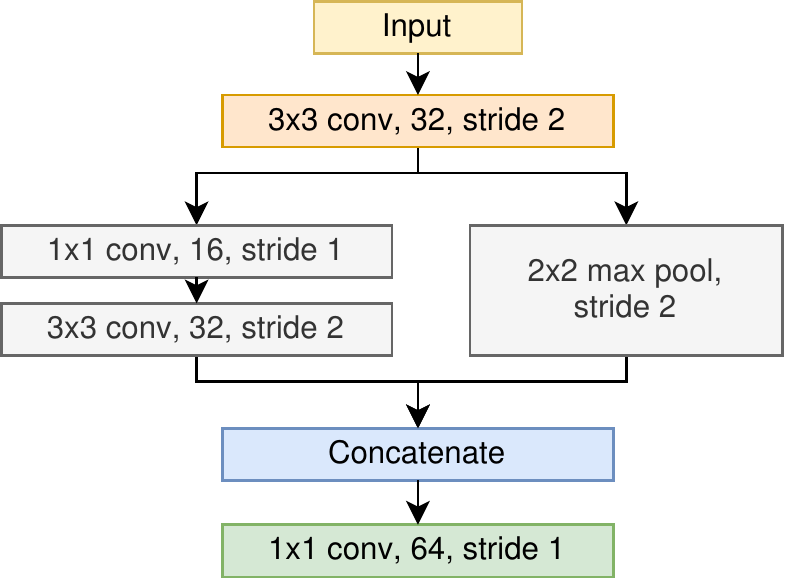}
    \caption{The architecture of STEM module.}
    \label{fig:stem}
    \vspace{-0.1in}
\end{figure}

\subsection{Help to Network Design}

daBNN is the first highly-optimized open-sourced BNN inference framework. It not only enables the deployment of BNNs in industries, but also brings help to BNN design for researchers. For example, it is a common practice that the first and last layer in BNNs remain full precision. However, no one find that the first layer, which is usually a convolution with large kernel, often takes up more than half latency of a binary neural network. Given this observation, we replace the first 7$\times$7 convolution layer in Bi-Real Net 18 \cite{liu2018bi} with a STEM module \cite{wang2018pelee} as shown in Figure \ref{fig:stem}, then we get a 30\% speedup on Google Pixel 1 without any accuracy loss as shown in Table \ref{tab:design_stem}. We cannot make this easy but effective improvement if we use an under-optimized framework like BMXNet, since the binary convolutions in it are even slower than float convolutions.

\section{Model Conversion Tool}

We present a model conversion tool, named onnx2bnn, to convert trained BNN models to our daBNN format. We provide pre-built onnx2bnn binaries for Linux, macOS and Windows, and thus no compilation is required for our users.

Our model conversion tool supports ONNX (Open Neural Network Exchange) \cite{onnx} format, which is greatly supported or officially integrated by many frameworks and tools \cite{tensorflow2015-whitepaper, paszke2017automatic, chen2015mxnet, seide2016cntk}. We only depend on the standard ONNX operators to ensure the interoperability. By contrast, BMXNet implements and depends on some custom MXNet operators (like ``QConvolution''), so the BNN models trained in other deep learning frameworks (\eg TensorFlow \cite{tensorflow2015-whitepaper} and PyTorch \cite{paszke2017automatic}) cannot be deployed on BMXNet easily.

Our model conversion tool recognizes whether a tensor is binary by various ways, \eg checking whether the tensor is an output of a Sign operator. The convolutions with binary input and binary weight are then converted to binary convolutions in daBNN format. The weight of binary convolutions will be packed so that the model size will be drastically compressed (32$\times$ if all weights are packed).

\begin{table}[]
    \centering
    \begin{tabular}{c|c|c|c}
        \hline
         &  \makecell{ImageNet \\ top-1} & \makecell{Latency of \\ Float OPs} & Latency \\
        \hline
        Bi-Real Net 18 & 56.4\% & 37.2ms & 61.3ms \\
        \hline
        \makecell{Bi-Real Net 18 \\ with STEM} & \textbf{56.4\%} & \textbf{20.1ms} & \textbf{43.2ms}    \\
        \hline
    \end{tabular}
    \caption{Based on our daBNN, simply replacing the 7$\times$7 full precision convolution layer with STEM module can brings 30\% speedup. It shows that our daBNN offers a fertile ground for designing more efficient BNN architectures.}
    \label{tab:design_stem}
    \vspace{-0.3in}
\end{table}

\section{Conclusion and Future Work}

We presented daBNN, a super fast binary neural networks inference framework for ARM devices. We implement binary convolution and other operators by ARM assembly. Extensive experiments show that our daBNN is extremely faster than both BMXNet and modern full precision inference frameworks. We believe that daBNN will greatly help both deployment and design of BNNs. For the ease of use, we publish pre-built binaries, libraries and also a sample Android project. Our source code, sample project, documentation and pre-trained models are published on GitHub. In the future, we are going to implement BNNs on X86 and RISC-V. We are also looking forward to cooperating with research teams to design or search better BNN structures.

{\small
\bibliographystyle{ACM-Reference-Format}
\bibliography{ref}
}

\end{document}